\documentclass[conference]{IEEEtran}
\IEEEoverridecommandlockouts

\def\BibTeX{{\rm B\kern-.05em{\sc i\kern-.025em b}\kern-.08em
		T\kern-.1667em\lower.7ex\hbox{E}\kern-.125emX}}
	
\usepackage{adjustbox}	
\usepackage{amsmath,amssymb,amsfonts}
\usepackage{algorithmic}
\usepackage{graphicx}
\usepackage[]{algorithm2e}
\usepackage{textcomp}
\usepackage{xcolor}
\usepackage{epstopdf}
\usepackage{multirow}
\usepackage{mathtools}
\usepackage{pbox}
\graphicspath{{Img/}}

\begin{document}
\title{Efficient single input-output layer spiking neural classifier with time-varying weight model}

\author{
	\IEEEauthorblockN{1\textsuperscript{st} Abeegithan Jeyasothy}
	\IEEEauthorblockA{\textit{School of Computer Science and Engineering} \\
		\textit{Nanyang Technological University}\\
		Singapore \\
		abee0001@e.ntu.edu.sg}
	\and
	\IEEEauthorblockN{2\textsuperscript{nd} Savitha Ramasamy}
	\IEEEauthorblockA{\textit{Institute for Infocomm Research} \\
		\textit{Agency for Science, Technology and Research}\\
		Singapore \\
		ramasamysa@i2r.a-star.edu.sg}
	\and
	\IEEEauthorblockN{3\textsuperscript{rd} Suresh Sundaram}
	\IEEEauthorblockA{\textit{Department of Aerospace Engineering} \\
		\textit{Indian Institute of Science, Bangalore }\\
		India\\
		sureshsundaram@iisc.ac.in}
	
}

\maketitle

\begin{abstract}
	This paper presents a supervised learning algorithm, namely, the Synaptic Efficacy Function with Meta-neuron based learning algorithm (SEF-M) for a spiking neural network with a time-varying weight model. For a given pattern, SEF-M uses the learning algorithm derived from meta-neuron based learning algorithm to determine the change in weights corresponding to each presynaptic spike times. The changes in weights modulate the amplitude of a Gaussian function centred at the same presynaptic spike times. The sum of amplitude modulated Gaussian functions represents the synaptic efficacy functions (or time-varying weight models). The performance of SEF-M is evaluated against state-of-the-art spiking neural network learning algorithms on 10 benchmark datasets from UCI machine learning repository. Performance studies show superior generalization ability of SEF-M. An ablation study on time-varying weight model is conducted using JAFFE dataset. The results of the ablation study indicate that using a time-varying weight model instead of single weight model improves the classification accuracy by $14\%$. Thus, it can be inferred that a single input-output layer spiking neural network with time-varying weight model is computationally more efficient than a multi-layer spiking neural network with long-term or short-term weight model.
\end{abstract}


\section{Introduction}

Spiking Neural Networks (SNNs) are proven to be computationally more powerful than sigmoidal neural networks~\cite{Masss_neuron}. However, there are challenges in developing supervised learning algorithms for SNN that completely exploit its computational power. Adaptation of the backpropagation to the realm of SNN~\cite{SpikeProp,Mutli_SpikeProp} is severely limited by the silent neuron problem that arises due to events of no spikes.

There have been several research efforts to develop supervised learning algorithms for SNN to optimize its computational power. Spike-Time-Dependent Plasticity (STDP)~\cite{STDP_overview} is one of the commonly used rules in supervised learning algorithms. STDP rule estimates the weight update by computing the difference in time of presynaptic spikes with respect to the postsynaptic spike. Due to the local characteristic of STDP rule, supervised learning algorithms use STDP rule in a normalized form or in conjunction with other learning mechanisms (Remote Supervised Method (ReSuMe)~\cite{ReSuMe}, Synaptic Weight Association Training (SWAT)~\cite{SWAT}, Accurate Synaptic-efficiency Adjustment (ASA)~\cite{ASA}, etc.). Unlike STDP rule, there are learning algorithms that use the error between the firing threshold and the postsynaptic potential~\cite{Tempotron} or the error between the spike responses~\cite{SPAN} to estimate the weight updates. There are also algorithms with rank order scheme~\cite{SRESN,SpikeTemp}, where the order of the spike is used to estimate the weight updates.

Although there are several types of synaptic plasticity (weight) models, most of these algorithms~\cite{SpikeProp, SPAN,SRESN,ReSuMe,Tempotron, Chronotron,OMLA,OSNN,MuSpiNN, Mutli_SpikeProp,ASA,SpikeTemp} have been developed for long-term plasticity model. Another type of plasticity model is the short-term plasticity model~\cite{1996LiawSynapseModel,1996MarkramdepressingSynapse,1997MarkramdepressingSynapse,1998MarkramSynapseModel,STP_Model2,STP_model1}. In a short-term plasticity model, weights are updated during spiking activity and recovers back to constant value. It is established in the literature~\cite{1996LiawSynapseModel, 1999MaassSynapseModel} that an SNN with short-term plasticity is computationally powerful than an SNN with long-term plasticity. However, they result in a large network with huge computational load and also important to note that short-term plasticity models are used in conjunction with long-term plasticity models.~\cite{1996LiawSynapseModel,2003LiawModelImprovement,1998LiawModelImprovement, 2001LiawmodelImprovement,SWAT}.

A new time-varying weight model based neuron, namely, Synaptic Efficacy Function based neuRON (SEFRON)~\cite{SEFRON}, has been proposed to exploit the computational power of spiking neuron with a dynamic synaptic plasticity model. In a SEFRON, the weight is a function represented as a summation of amplitude modulated Gaussian functions. A single SEFRON classifier 
has no hidden layers and only one output neuron to solve binary classification problems. It is observable from~\cite{SEFRON} that the SEFRON classifier performs comparably well to an SNN classifier with many layers and neurons in solving binary classification tasks.

In this paper, we develop a multi-class SEFRON classifier with meta-neuron based learning rule (SEF-M). The multi-class SEFRON classifier has multiple output neurons corresponding to the number of classes with no hidden layers. Each input-output connection is a time-varying weight function. The learning rule of the SEF-M is derived based on the meta-neuron learning rule proposed in~\cite{OMLA} that uses a meta-neuron as an alternative to the STDP learning rule. For a given pattern, meta-neuron determines the weight sensitivity modulation factor for each synapse from the collective firing of all the synapses. Weight sensitivity modulation factor is then used to determine the individual change in weight for each synapse. While STDP rule uses a learning window, meta-neuron learning rule entirely relies on spiking activities to determine the change in weight. This makes it a suitable candidate to handle data with multiple clusters for each class. In the meta-neuron based learning rule, weight sensitivity modulation factor is determined for each synapse as it uses long-term synaptic plasticity model (constant weight). In this paper, the meta-neuron based learning rule is modified such that the weight sensitivity modulation factor is determined for each presynaptic spikes instead of each synapse. Weight update corresponding to each presynaptic spikes are determined by the new weight sensitivity modulation factor. The time-varying weight function is obtained by adding Gaussian functions centred at each presynaptic spikes and the amplitude of the function is modulated by the amplitude of the change in weight corresponding to those presynaptic spikes. 

The performance of the SEF-M is compared with those of SpikeProp~\cite{SpikeProp}, SWAT~\cite{SWAT}, Two stage Margin Maximization Spiking Neural Network (TMM-SNN)~\cite{TMM-SNN} and SEFRON~\cite{SEFRON} on ten benchmark classification problems from the University of California, Irvine California (UCI) machine learning repository~\cite{UCI}. It can be observed from the performance results that the SEF-M outperforms the other algorithms, despite having a simple architecture. This can be attributed to the time-varying nature of the weight function in SEF-M. We also conducted an ablation study on the time-varying weight function using the Japanese Female Facial Expression (JAFFE) dataset. The results of the study indicate that using a time-varying weight model instead of a single weight (long-term weight) model improves the classification accuracy. 

The remaining of the paper is organized as follows: We present the preliminaries of SEFRON and meta-neuron based learning algorithm, and then present the learning algorithm of SEF-M in section~\ref{section:method}. Section~\ref{section:classification} briefly summarizes the training methodology of the SEF-M for solving multi-class classification problems. We present the performance results of the SEF-M, in comparison with other state-of-the-art algorithms in Section~\ref{section:results}. We present the ablation study of time-varying weight model in Section~\ref{section:jaffe} and summarize the overall study in Section~\ref{section:conclusion}.

\section{Methods}
\label{section:method}

In this section, we first present the preliminaries of the SEFRON and the meta-neuron based learning algorithm and then present the architecture and learning algorithm of a SEF-M. 

\subsection{SEFRON}

A binary class SEFRON classifier uses single output neuron with a time-varying weight model. In SEFRON classifier, the weight between an input and the output neuron is a time-varying weight function referred to as synaptic efficacy function. This synaptic efficacy function is obtained by summing amplitude modulated Gaussian function centred at different presynaptic spike times. The centre and the amplitude of the Gaussian functions are determined by selected presynaptic spike times and the momentary weight change corresponding those presynaptic spikes. 

For a given input pattern, any learning rule for SEFRON should determine the momentary weight change corresponding to the selected presynaptic spike. The final weights are expected to be similar to the other patterns that are similar to the given input pattern. Hence, the momentary weight change is modulated by a Gaussian function to produce weights that are similar to the current one if the presynaptic spikes are nearer. 

The final synaptic efficacy function may have both positive and negative values within the interval as a consequence of both positive and negative weight updates. For reader’s deeper perusal, detailed derivation of supervised learning rule proposed for binary-class SEFRON classifier can be found in~\cite{SEFRON}.

\subsection{Meta-neuron based learning algorithm}

Meta-neuron based Learning Algorithm is a supervised learning algorithm for SNN proposed in~\cite{OMLA}. This is implemented for an online learning (one-shot learning) environment that uses single layer (input-output) architecture with evolving output neurons and constant weights (long-term plasticity model). For a given pattern, the weight of the meta-neuron corresponding to each synapse is calculated from the difference between the normalized postsynaptic potential and the weight of the synapse.  The weight of meta-neuron is used to determine the weight sensitivity modulation factor for each synapse. The sum of the weight sensitivity modulation factor across all the synapse is equal to $1$. Weight update in each synapse occurs to minimize the error between the postsynaptic potential and the firing threshold of the postsynaptic neuron. Weight sensitivity modulation factor determines the amplitude of the change in weight for each synapse from the error.  

In~\cite{OMLA}, the change in weight is determined for each synapse, not for each presynaptic spikes. Hence the learning rule in~\cite{OMLA} cannot be directly applied for a multi-class SEFRON. It has to be modified such that the change in weight corresponding to each presynaptic spikes can be determined in order to train the classifier with a time-varying weight model. 

\subsection{Synaptic Efficacy Function with Meta-neuron based learning algorithm (SEF-M)}

In this section, we present the architecture and learning algorithm of SEF-M. First, we present the architecture of SEF-M in section~\ref{section:architecture} and then the learning algorithm in section~\ref{section:learning_algo}. 

\subsubsection{Architecture}
\label{section:architecture}

\begin{figure}[htb]	
	\centering
	\includegraphics[width=0.8\linewidth]{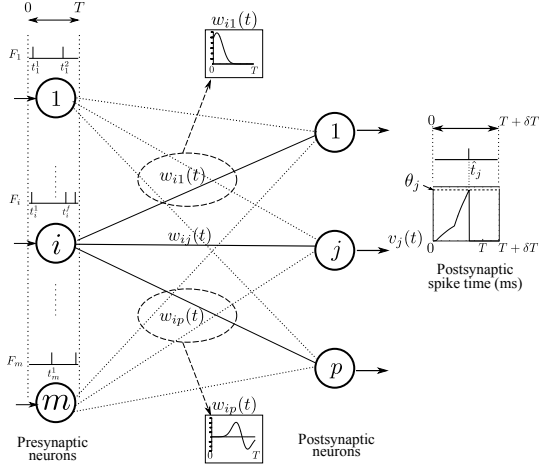}
	\caption{A SEF-M classifier with $m$ number of input neurons and $p$ number of output neurons. Presynaptic spike time for a given input pattern is in the interval of $[0, T]ms$ and postsynaptic spike time is in the interval of $[0, T+\delta T]ms$. Weight of each synapse (eg. $w_{i1}(t),w_{ip}(t)$) is a synaptic efficacy function within the interval of $[0, T]ms$. }
	\label{fig:Sefron_model}
\end{figure}

The SEF-M classifier is a spiking neural network with 2 layers namely, the input and output layer. Fig~\ref{fig:Sefron_model} shows a SEF-M classifier with $m$ number of input neurons and $p$ number of output neurons. Each neuron in the input layer represents an input feature, while each neuron in the output layer represents a class. It must be noted that the SEF-M classifier does not have a hidden layer, and there are direct connections between the input and output neurons. Each input-output connection is a synaptic efficacy function that represents a time-varying weight model. Each input neuron fires a set of presynaptic spikes. The set of presynaptic spike times $F_i$ of the $i^{th}$ synapse is defined as,
\begin{equation}
F_i= \{t^k_i; 1 \leq k \leq n_i\}
\end{equation}  
where $n_i$ is the total number of presynaptic spikes fired by the $i^{th}$ synapse and $k$ represents the order of the firing. Firing time of $k^{th}$ presynaptic spike fired by the $i^{th}$ input neuron is denoted as $t^k_i$. 

The postsynaptic potential of the $j^{th}$ postsynaptic neuron $v_j(t)$ is defined as the summation of the product of spike response function and momentary weight. 
\begin{equation}
v_j(t)=\sum_{i=1}^{m}\sum_{k=1}^{n_i} w_{ij}(t^k_i).\epsilon(t-t^k_i)
\end{equation}

The momentary weight $w_{ij}(t^k_i)$ is obtained by sampling the synaptic efficacy function $w_{ij}(t)$ at time $t^k_i$. The spike response function $\epsilon(t)$ is expressed as, 
\begin{equation}
\epsilon(t) = \frac{t}{\tau}exp(1-\frac{t}{\tau}).H(t)
\label{equation:Spike_response}
\end{equation}
here $\tau$ is the time constant and $H(t)$ is a Heaviside step function.  $j^{th}$ postsynaptic neuron fires a postsynaptic spike when the $v_j(t)$ crosses the threshold $\theta_j$.

\subsubsection{Learning algorithm}
\label{section:learning_algo}

Meta-neuron based learning rule updates the weights to minimize the error ($\Delta v_j$) between the postsynaptic potential and the firing threshold. The error $\Delta v_j$ is calculated as,
\begin{equation}
\Delta v_j=\theta_j - v_j(\hat{t})
\label{equation:PSP_error}
\end{equation}
where $v_j(\hat{t})$ is the postsynaptic potential at the reference postsynaptic time $\hat{t}$.

Meta-neuron based learning rule determines the individual change in weight for each synapse connected to output neuron $j$ to minimize the overall error $\Delta v_j$. This rule is changed for SEFRON model as,  the individual change in weight corresponding to each presynaptic spikes (instead of each synapse) should minimize the overall error $\Delta v_j$. This new rule can be expressed as, 
\begin{equation}
\sum_{i=1}^{m}\sum_{k=1}^{n_i} \Delta w_{ij}(t^k_i).\epsilon(\hat{t}-t^k_i)=\Delta v_j
\label{equation:PSP_change}
\end{equation}

For each presynaptic spike $t^k_i$, weight of meta-neuron $z^k_{ij}$ is calculated as, 
\begin{equation}
z^k_{ij}=\begin{cases}
u^k_{ij}(\hat{t})-w_{ij}(t^k_i) &\text{if $u^k_{ij}(\hat{t})>w_{ij}(t^k_i)$} \\
o &\text{otherwise}
\end{cases}
\end{equation}
here $u^k_{ij}(\hat{t})$ is the normalized postsynaptic potential at $\hat{t}$ due to presynaptic spike at $t^k_i$. It is calculated as, 
\begin{equation}
u^k_{ij}(\hat{t})=\frac{\epsilon(\hat{t}-t^k_i)}{\sum_{i=1}^{m}\sum_{k=1}^{n_i}\epsilon(\hat{t}-t^k_i)}
\end{equation}

$M^k_{ij}$ is the weight sensitivity modulation factor for presynaptic spike at $t^k_i$. It is calculated using $z^k_{ij}$ and expressed as, 
\begin{equation}
M^k_{ij}=\frac{z^k_{ij}.\epsilon(\hat{t}-t^k_i)}{\sum_{i=1}^{m}\sum_{k=1}^{n_i}z^k_{ij}.\epsilon(\hat{t}-t^k_i)}
\end{equation}
$M^k_{ij}$ can be interpreted as the ratio of postsynaptic potential of the meta-neuron at $\hat{t}$ induced by the presynaptic spike at $t^k_i$ and the total postsynaptic potential of the meta-neuron at $\hat{t}$. Hence the summation of $M^k_{ij}$ for all the presynaptic spikes fired by all the synapse is equal to $1$.

Individual momentary change in weight corresponding to each presynaptic spike $\Delta w_{ij}(t^k_i)$ in equation~\ref{equation:PSP_change} is calculated as, 
\begin{equation}
\Delta w_{ij}(t^k_i) = M^k_{ij} \frac{\Delta v_j}{\epsilon(\hat{t}-t^k_i)}
\label{equation:momentary_weight}
\end{equation}

Equations~\ref{equation:PSP_error} to~\ref{equation:momentary_weight} are derived from meta-neuron based learning rule. Here the focus is only on the important modifications needed in the meta-neuron based learning rule~\cite{OMLA} to make it suitable to train the weights of the time-varying weight model in SEFRON.

The momentary weight $\Delta w_{ij}(t^k_i)$ calculated in equation~\ref{equation:momentary_weight} is embedded in a time interval by a Gaussian function $g_{ij}^k(t)$ as,

\begin{equation}
g_{ij}^k(t) = \Delta w_{ij}(t^k_i).exp(\frac{-(t-t^k_i)^2}{2\sigma^2})
\label{equation:efficacy_update_spike}
\end{equation}
here $\sigma$ is the efficacy update range. A larger value for $\sigma$ result in a constant function which is similar to a long-term synaptic plasticity model. 

Synaptic efficacy function $w_{ij}(t)$ (time-varying weight function) for each synapse is obtained by adding the $g_{ij}^k(t)$ calculated for all the presynaptic spikes fired by the same synapse. The new synaptic efficacy function $w_\text{ij new}(t) $ is updated as,
\begin{equation}
w_\text{ij new}(t) = w_\text{ij old}(t) + \lambda.\sum_{k=1}^{n_i}g_{ij}^k(t)
\label{equation:Synaptic_efficacy_update}
\end{equation}

\section{SEF-M for a classification problem}
\label{section:classification}

SEF-M classifier uses the parameter update strategy derived from~\cite{OMLA} and also uses the skip sample strategy in~\cite{SEFRON} during the training process.

\subsubsection{Parameter update strategy}

In a supervised learning framework,  for a given input pattern a reference signal (coded output class label) is used for each output neuron to determine the error and update the weights associated with that output neuron. Here a postsynaptic spike time $\hat{t}$ is used as the reference signal. For a given pattern, the parameter update strategy is used to determine the reference signal for all the postsynaptic neurons. 

Parameter update strategy is employed to train the correct class postsynaptic neuron to fire at desired firing time $\hat{t}_d$ and the wrong class postsynaptic neurons to fire further away from $\hat{t}_d$. Synaptic efficacy function $w(t)$ of the synapse connected to correct class postsynaptic neuron is updated only when the actual firing time $\hat{t}^c_a$ of the correct class postsynaptic neuron is greater than $T_d$. $T_d$ is a heuristic criterion introduced in~\cite{OMLA} to avoid over-fitting and calculated as,
\begin{equation}
T_d= \hat{t}_d+ \alpha_d.(T-\hat{t}_d)
\end{equation}
here $T$ is the presynaptic spike interval. $\alpha_d$ is referred to as delete threshold, it can be interpreted as the allowance for the similarity between patterns that reduces over-fitting.

The reference firing time $\hat{t}^c_r$ to update the $w(t)$ of the correct class postsynaptic neuron is calculated as,
\begin{equation}
\hat{t}^c_r= (1- \alpha_s).\hat{t}^c_a
\end{equation}
$\alpha_s$ is the learning rate for the reference firing time $\hat{t}^c_r$. Every update moves the reference firing time closer to the desired firing time.

A margin time ($T_m$) is the minimum time allowed between the firing time of a correct and wrong class post-synaptic neuron.  $T_m$ is calculated as, 
\begin{equation}
T_m=\alpha_m.(T-\hat{t}_d)
\end{equation}
here $\alpha_m$ is referred to as margin threshold.

$w(t)$ of the wrong class postsynaptic neurons are updated in order to fire a postsynaptic spike after the desired firing time with at least a delay of $T_m$. Reference firing time $\hat{t}^w_r$ for wrong class postsynaptic neuron is calculated for two cases as given below,
\begin{equation}
\hat{t}^w_r= \begin{cases}
\hat{t}^c_a +T_m & \hat{t}^c_a \leq T_d\\
\hat{t}^c_r +T_m & \hat{t}^c_a > T_d\\
\end{cases}
\end{equation}

An input sample is not used to update (skip sample strategy) the $w(t)$ of any synapse only if the correct postsynaptic neuron fires before $T_d$ and the margin time requirements for the wrong postsynaptic neuron are satisfied. Otherwise $w(t)$ of all the synapses are updated using equations~\ref{equation:PSP_error} to~\ref{equation:Synaptic_efficacy_update} with their respective reference firing time ($\hat{t}=\hat{t}^c_r$ or $\hat{t}=\hat{t}^w_r$).

\subsubsection{Initialization}

The first sample presented from each class is used to initialize the firing threshold and synaptic efficacy function. 
\begin{equation}
\theta_j := \sum_{i=1}^{m}\sum_{k=1}^{n_i}u^k_{ij}(\hat{t}_d).\epsilon(\hat{t}_d-t^k_i)
\label{equation:Initial_theta}
\end{equation}

\begin{equation}
w_\text{ij initial}(t) = \sum_{k=1}^{n_i}u^k_{ij}(\hat{t}_d).exp(\frac{-(t-t^k_i)^2}{2\sigma^2})
\label{equation:Initial_W}
\end{equation}

The pseudo code for SEF-M learning rule is given in algorithm~\ref{Algo:Pseudo_code}. 

\begin{algorithm}[htb]
	\caption{Pseudo code to train a SEF-M classifier}
	\label{Algo:Pseudo_code}
	\For{all samples}{
		\eIf{first sample from any class}{
			Initialize $w(t)$ of all the synapses and $\theta$ of the corresponding postsynaptic neuron
			(equation~\ref{equation:Initial_W} and \ref{equation:Initial_theta} respectively)\;
		}
		{
			$\hat{t}^c_a \gets$ postsynaptic firing time of correct class\;
			$\hat{t}^w_a \gets$ postsynaptic firing times of wrong classes\;
			
			\eIf{$\hat{t}^c_a \leq T_d$}			
			{\eIf{$\hat{t}^w_a-\hat{t}^c_a<T_m$}{   
					Update $w(t)$ of all the wrong class neruon that satisfy the above condition with $\hat{t}=\hat{t}^w_r$ (case1);
				}
				{
					The sample is not used to update any $w(t)$;
				}
				
			}	
			{Update $w(t)$ of the correct class neuron with $\hat{t}=\hat{t}^c_r$
				
				\If{$\hat{t}^w_a-\hat{t}^c_r<T_m$}{ 
					Update $w(t)$ of all the wrong class neruon that satisfy the above condition with $\hat{t}=\hat{t}^w_r$ (case2);
				}
			}	
			
		}	
	}
\end{algorithm}

\section{Experimental results for UCI machine learning dataset}
\label{section:results}
Performance of SEF-M classifier is evaluated on ten benchmark datasets from UCI machine learning repository and compared with other existing algorithms. For all the experiments the real-valued input data is converted into spike patterns using population encoding scheme~\cite{SRESN,SpikeProp}. To enable fairness in evaluation, we use $6$ receptive field neurons and overlap constant set to $0.7$ in population encoding. The learning rate for the reference firing time $\alpha_s$ and the efficacy update range $\sigma$ are the only two problem-dependent parameters and were set by trial and error by cross-validating to get the highest performance.  All the other parameters are selected based on the guidelines given in~\cite{OMLA} and set as follows, 
\begin{itemize}
	\item Presynaptic spike interval limit $T$ -$3ms$
	\item Postsynaptic spike interval - $[0,8]ms$
	\item Time constant $\tau$ of spike response function- $3ms$
	\item Desired firing time $\hat{t}_d$ -$2ms$
	\item learning rate $\lambda$ - $0.1$
	\item margin threshold $\alpha_m$ -$0.3$
	\item delete threshold $\alpha_d$-$0.25$
\end{itemize}

\begin{table}[htb]
	\centering
	\caption{Description of Dataset used for validation}
	\label{table:datasets}
	\resizebox{\columnwidth}{!}{
		\begin{tabular}{|l|c|c|l|l|}
			\hline
			\multicolumn{1}{|c|}{\multirow{2}{*}{Dataset}} & \multicolumn{1}{c|}{\multirow{2}{*}{\# Features}} & \multicolumn{1}{c|}{\multirow{2}{*}{\# Classes}} & \multicolumn{2}{l|}{\# Samples} \\ \cline{4-5} 
			\multicolumn{1}{|c|}{}                         & \multicolumn{1}{c|}{}                             & \multicolumn{1}{c|}{}                            & Training        & Testing       \\ \hline
			
			Iris            & 4               & 3          & 75            & 75          \\ \hline
			Wine       & 13              & 3          & 60            & 118           \\ \hline
			Acoustic emission           & 5               & 4          & 62            & 137           \\ \hline
			Breast Cancer          & 9              & 2          & 350            & 333           \\ \hline
			Echo-cardiogram          & 10              & 2          & 66           & 65          \\ \hline
			Mammogram         & 9              & 2          & 80           & 11           \\ \hline
			Liver          & 6              & 2          & 170            & 175           \\ \hline
			PIMA          & 8              & 2          & 384            & 384           \\ \hline
			Ionosphere         & 34              & 2          & 175            & 176          \\ \hline
			Hepatitis          & 19              & 2          & 78            & 77          \\ \hline
		\end{tabular}
	}
\end{table}

Description of the ten UCI machine learning dataset and the split of the training and the testing dataset is given in table~\ref{table:datasets}. For each dataset, 10-random fold validation is conducted. Performance of SEF-M is compared with SpikeProp~\cite{SpikeProp}, SWAT~\cite{SWAT},  TMM-SNN~\cite{TMM-SNN}, and SEFRON~\cite{SEFRON} using three metrics; viz the architecture of the network, the training, and the testing accuracies. Architecture is in the form of $N_i-N_h-N_j$, where $N_i,N_h$ and $N_j$ are the total number of input, hidden and output neurons respectively. For evolving architecture, the range for the number of neurons is given. Experiments for SpikeProp~\cite{SpikeProp}, SWAT~\cite{SWAT}, and TMM-SNN~\cite{TMM-SNN} used for comparison are not conducted and the results are reproduced from the literature~\cite{TMM-SNN}. For SEFRON~\cite{SEFRON}, only experiments for echo-cardiogram, mammogram and hepatitis are conducted. Parameters for those experiments are set based on cross-validation and the guidelines given in~\cite{SEFRON}. Remaining binary-class results are reproduced from the literature~\cite{SEFRON}. It can be noted that SEFRON is a binary-class classifier, hence the experiments for iris, wine, and acoustic emission dataset cannot be conducted. Table~\ref{table:performance} shows the performance comparison of SEF-M with other algorithms.

\begin{table*}	
	\caption{Performance comparison on UCI datasets}
	\label{table:performance}
	\centering
	\begin{adjustbox}{totalheight=8in}
		\begin{tabular}{|c|c|c|c|c|}
			
			\hline
			Dataset                                 & Algorithm                                                                               & Architecture                                                                                        & \begin{tabular}[c]{@{}c@{}}Training \\ Accuracy (\%)\end{tabular}                                 & \begin{tabular}[c]{@{}c@{}}Testing\\ Accuracy (\%)\end{tabular}                                    \\ \hline
			\multicolumn{1}{|c|}{Iris}              & \begin{tabular}[c]{@{}c@{}}SpikeProp\\ SWAT\\ TMM-SNN\\ SEF-M\end{tabular} & \begin{tabular}[c]{@{}c@{}}25-10-3\\ 24-312-3\\  24-(4-7)-3\\ \textbf{24-3}\end{tabular}          & \begin{tabular}[c]{@{}c@{}}97.2(1.9)\\ 96.7(1.4)\\ 97.5(0.8)\\ 98.0(1.7)\end{tabular} & \begin{tabular}[c]{@{}c@{}}96.7(1.6)\\ 92.4(1.7)\\  97.2(1.0)\\ \textbf{97.6(1.5)}\end{tabular}  \\ \hline
			\multicolumn{1}{|c|}{Wine}              & \begin{tabular}[c]{@{}c@{}}SpikeProp\\ SWAT\\ \ TMM-SNN\\ SEF-M\end{tabular} & \begin{tabular}[c]{@{}c@{}}79-10-3\\ 78-1014-3\\  78-3-3\\ \textbf{78-3}\end{tabular}             & \begin{tabular}[c]{@{}c@{}}99.2(1.2)\\ 98.6(1.1)\\  100(0)\\ 100(0)\end{tabular}       & \begin{tabular}[c]{@{}c@{}}96.8(1.6)\\ 92.3(2.4)\\  97.5(0.8)\\ \textbf{97.8(1.1)}\end{tabular}  \\ \hline
			\multicolumn{1}{|c|}{Acoustic Emission} & \begin{tabular}[c]{@{}c@{}}SpikeProp\\ SWAT\\ TMM-SNN\\ SEF-M\end{tabular} & \begin{tabular}[c]{@{}c@{}}31-10-4\\ 30-390-4\\ 30-(4-7)-4\\ \textbf{30-4}\end{tabular}           & \begin{tabular}[c]{@{}c@{}}98.5(1.7)\\ 93.1(2.3)\\  97.6(1.3)\\ 99.2(0.9)\end{tabular} & \begin{tabular}[c]{@{}c@{}}97.2(3.5)\\ 91.5(2.3)\\  97.5(0.7)\\ \textbf{99.1(0.3)}\end{tabular}  \\ \hline
			Breast Cancer                           & \begin{tabular}[c]{@{}c@{}}SpikeProp\\ SWAT\\  TMM-SNN\\ SEFRON\\SEF-M\end{tabular} & \begin{tabular}[c]{@{}c@{}}55-15-2\\ 54-702-2\\  54-(2-8)-2\\55-1\\ \textbf{54-2}\end{tabular}          & \begin{tabular}[c]{@{}c@{}}97.3(0.6)\\ 96.5(0.5)\\ 97.4(0.3)\\98.3(0.8)\\ 98.2(0.8)\end{tabular} & \begin{tabular}[c]{@{}c@{}}97.2(0.6)\\ 95.8(1.0)\\  97.2(0.5)\\96.4(0.7)\\ \textbf{98.0(0.4)}\end{tabular}  \\ \hline
			Echo-cardiogram                          & \begin{tabular}[c]{@{}c@{}}SpikeProp\\ SWAT\\ TMM-SNN\\ SEFRON\\SEF-M\end{tabular} & \begin{tabular}[c]{@{}c@{}}61-10-2\\ 60-780-2\\  60-(2-3)-2\\61-1\\ \textbf{60-2}\end{tabular}           & \begin{tabular}[c]{@{}c@{}}86.6(2.5)\\ 90.6(1.8)\\  86.5(2.1)\\ 88.6(3.9)\\86.5(9.2)\end{tabular} & \begin{tabular}[c]{@{}c@{}}84.5(3.0)\\ 81.8(2.8)\\  85.4(1.7)\\\textbf{86.5(3.3)}\\ 85.5(1.5)\end{tabular}  \\ \hline
			Mammogram                               & \begin{tabular}[c]{@{}c@{}}SpikeProp\\ SWAT\\  TMM-SNN\\ SEFRON\\ SEF-M\end{tabular} & \begin{tabular}[c]{@{}c@{}}55-10-2\\ 54-702-2\\ 54-(5-7)-2\\55-1\\ \textbf{54-2}\end{tabular}           & \begin{tabular}[c]{@{}c@{}}82.8(4.7)\\ 82.6(2.1)\\  87.2(4.4)\\92.8(5)\\ 88.1(4.5)\end{tabular} & \begin{tabular}[c]{@{}c@{}}81.8(6.1)\\ 78.2(12.3)\\  84.9(8.6)\\82.7(10)\\ \textbf{93.6(6.1)}\end{tabular} \\ \hline
			Liver                                   & \begin{tabular}[c]{@{}c@{}}SpikeProp\\ SWAT\\ TMM-SNN\\SEFRON\\ SEF-M\end{tabular} & \begin{tabular}[c]{@{}c@{}}37-15-2\\ 36-468-2\\ 36-(5-8)-2\\37-1\\ \textbf{36-2}\end{tabular}           & \begin{tabular}[c]{@{}c@{}}71.5(5.2)\\ 74.8(2.1)\\  74.2(3.5)\\91.5(5.4)\\ 74.8(3.9)\end{tabular} & \begin{tabular}[c]{@{}c@{}}65.1(4.7)\\ 60.9(3.2)\\ 70.4(2.0)\\ 67.7(1.3)\\\textbf{71.1(3.2)}\end{tabular}  \\ \hline
			PIMA                                    & \begin{tabular}[c]{@{}c@{}}SpikeProp\\ SWAT\\ TMM-SNN\\SEFRON\\ SEF-M\end{tabular} & \begin{tabular}[c]{@{}c@{}}49-20-2\\ 48-702-2\\ 48-(5-14)-2\\49-1\\ \textbf{48-2}\end{tabular}         & \begin{tabular}[c]{@{}c@{}}78.6(2.5)\\ 77.0(2.1)\\ 79.7(2.3)\\84.1(1.5)\\ 78.4(2.5)\end{tabular} & \begin{tabular}[c]{@{}c@{}}76.2(1.8)\\ 72.1(1.8)\\ \textbf{78.1(1.7)}\\74.0(1.2)\\ 77.3(1.3)\end{tabular}   \\ \hline
			Ionosphere                              & \begin{tabular}[c]{@{}c@{}}SpikeProp\\ SWAT\\ TMM-SNN\\ SEFRON\\SEF-M\end{tabular} & \begin{tabular}[c]{@{}c@{}}205-25-2\\ 204-2652-2\\  204-(23-34)-2\\205-1\\ \textbf{204-2}\end{tabular} & \begin{tabular}[c]{@{}c@{}}89.0(7.9)\\ 86.5(6.7)\\  98.7(0.4)\\97.0(2.5)\\ 98.3(2.6)\end{tabular} & \begin{tabular}[c]{@{}c@{}}86.5(7.2)\\ 90.0(2.3)\\ 92.4(1.8)\\88.9(1.7)\\ \textbf{93.2(1.8)}\end{tabular}  \\ \hline
			Hepatitis                               & \begin{tabular}[c]{@{}c@{}}SpikeProp\\ SWAT\\  TMM-SNN\\ SEFRON\\SEF-M\end{tabular} & \begin{tabular}[c]{@{}c@{}}115-15-2\\ 114-1482-2\\ 114-(3-9)-2\\ 115-1\\\textbf{114-2}\end{tabular}    & \begin{tabular}[c]{@{}c@{}}87.8(5.0)\\ 86.0(2.1)\\  91.2(2.5)\\94.6(3.5)\\ 90.9(5.7)\end{tabular} & \begin{tabular}[c]{@{}c@{}}83.5(2.5)\\ 83.1(2.2)\\  86.6(2.2)\\82.7(3.3)\\ \textbf{88.7(2.0)}\end{tabular}  \\ \hline
			
		\end{tabular}
	\end{adjustbox}
\end{table*}
From table~\ref{table:performance}, it can be seen that except for the PIMA and echo-cardiogram datasets, SEF-M outperforms all the algorithms in the remaining datasets for testing accuracy. For iris, wine, acoustic emission, breast cancer and echo-cardiogram datasets performance of all the algorithms are relatively well compared to the performance on other datasets. However, SEF-M uses the most simplest architecture with no hidden layers and fixed output neurons. The mammogram is a binary class dataset with high interclass overlaps. Testing accuracy for mammogram dataset is $8\%$ higher than the next best performing algorithm. It can also be noted that the training accuracy is lower than the testing accuracy for the mammogram dataset, a similar observation can be made for SWAT's performance on ionosphere dataset. The liver is another dataset that is not easily separable. Testing accuracy of SEF-M is $1\%$ higher than TMM-SNN. Similar observations in the performance can be made for ionosphere and hepatitis datasets. 

From the observations made, it can be highlighted that the SEF-M uses the most compact architecture. The results for the testing accuracy highlights that the performance of SEF-M is comparable for PIMA and echo-cardiogram datasets and better than all the other algorithms for the remaining datasets.

\subsection{Statistical Analysis of Performance Comparison}
The results of the performance comparison between SEF-M and other classifiers have been analysed by using the Friedman test followed by a pairwise comparison using the Fisher's LSD method as in~\cite{TMM-SNN}. The Friedman test was conducted with the null hypothesis that the performance of all the classifiers do not differ significantly. The null hypothesis is rejected with $95\%$ confidence interval if the p-value for the computed F-statistic is lower than $0.05$. The Friedman test is conducted using the mean testing accuracies on the ten datasets described in table~\ref{table:performance}. The SEFRON~\cite{SEFRON} is a binary-class classifier, therefore the results for multi-class datasets are not available. Hence, the statistical tests are conducted twice, first test is conducted without SEFRON's results and the second test is conducted only on the results for binary-class datasets. For the first Friedman test, a p-value of $4.8$e-$6$ is obtained. Hence, the null hypothesis can be rejected with a $95\%$ confidence interval. This indicates that the perform of all the algorithms are not same. Therefore, to analyse further, a pairwise comparison was performed using the Fisher's LSD method. The p-values of $6.8$e-$4$, $1.1$e-$6$ and $0.139$ are obtained for comparing SEF-M with SpikeProp, SWAT and TMM-SNN respectively. This indicates that SEF-M performs better than SpikeProp and SWAT with a $99\%$ confidence interval and performs better than TMM-SNN with a $85\%$ confidence interval. For the second Friedman test, a p-value of $6.6$e-$4$ is obtained. A p-value of $0.01$ is obtained for the pairwise comparison of SEF-M and SEFRON. This implies the SEF-M performs better then SEFRON with a $99\%$ confidence interval. 

These statistical results highlight that SEF-M performs better or equally well to other existing SNN algorithms on classification tasks.   

\section{Ablation study of time-varying weight model using JAFFE dataset}
\label{section:jaffe}
Japanese Female Facial Expression (JAFFE) dataset~\cite{Jaffe} contains 213 images of 10 females with 6 emotions (Anger, Happy, Dislike, Fear, Sad and Surprise) and one neutral expression. In total, JAFFE dataset has $7$ classes. Size of the original image is $256\times256$ pixels. For our experiment, the faces are cropped and resized to $100\times100$ pixels such that the eyes of all the images are aligned. We have conducted leave-one-out experiments to test the performance of SEF-M on JAFFE dataset. In leave-one-out experiments, one image from all the emotions for all the females are used as test dataset (total of 70 images) and the remaining images are used as training dataset (total of 143 images). Experiments are repeated 10 times with randomly selected train and test samples. Each pixel is considered as a feature, resulting in $10000$ features for an input image. The real-valued inputs from the images are converted into spike patterns using the population encoding scheme with $6$ receptive field neurons. Hence for this experiment, there are $60000$ ($6\times10000$) input neurons and $7$ output neurons in the architecture (no hidden layers). The other parameter settings are same as in section~\ref{section:results}. 
\begin{figure}[htb]	
	\centering
	\includegraphics[width=\linewidth]{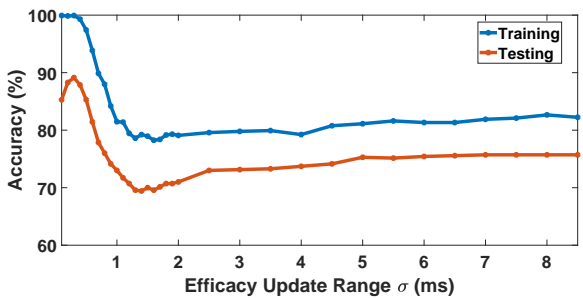}
	\caption{Performance of SEF-M on JAFFE dataset for different $\sigma$ values. }
	\label{fig:Sigma}
\end{figure}

Dynamic nature of the time-varying weight function is determined by the efficacy update range ($\sigma$). Hence influence of $\sigma$ in the SEF-M classifier is analysed by varying the value of $\sigma$ and conducting the experiments. Value of $\sigma$ is increased in steps of $0.1s$ from $0.1s$ to $2s$ and increased in steps of $0.5s$ until $8s$. For the final set of experiments, the value of $\sigma$  is set to infinity. Fig~\ref{fig:Sigma} shows the classification accuracy for different $\sigma$ values. The accuracies in Fig~\ref{fig:Sigma} are the mean accuracy value of the 10 random trials. It can be observed in Fig~\ref{fig:Sigma}, that the accuracy is high for the lower value of $\sigma$ and drops significantly when the value of $\sigma$ increases. Beyond $\sigma>2.5ms$ the testing accuracy is nearly constant. 

From table~\ref{table:performance_jaffe}, it can be seen that training accuracy of $99.93\%$ and testing accuracy of $89.14\%$ are obtained when $\sigma=0.3ms$. Training and testing accuracies are $82.24\%$ and $75.71\%$ respectively when the $\sigma$ is set to infinity. Setting the value of $\sigma$ to infinity is the same as using a constant weight model. A drop of $17.7\%$ and $13.43\%$ are observed for training and testing accuracies respectively if single weight model is used instead of time-varying weight model. This highlights that using time-varying weight model improves the classification accuracy for a given learning algorithm. 

\begin{table}[htb]
	\caption{Performance comparison on JAFFE dataset}
	\label{table:performance_jaffe}
	\centering
	\begin{tabular}{|l|l|l|}
		\hline
		Method & \begin{tabular}[c]{@{}l@{}}Training\\ Accuracy(\%)\end{tabular} & \begin{tabular}[c]{@{}l@{}}Testing\\ Accuracy(\%)\end{tabular} \\ \hline
		SEF-M (with single weight) & 82.24 & 75.71 \\ \hline
		SEF-M (with time-varying weight) & 99.93 & 89.14 \\ \hline
	\end{tabular}
\end{table}

\section{Conclusion}
\label{section:conclusion}
In this paper, a supervised learning rule referred to as SEF-M for multi-class classification problem has been presented. SEF-M rule has been developed for a single input-output layer spiking neural network classifier with the time-varying weight model. SEF-M uses meta-neuron based learning rule to determine the change in weight and uses Gaussian distribution function to obtain the time-varying synaptic efficacy function.  

The performance of SEF-M classifier is compared with four other well known SNN classifiers in literature for ten benchmark datasets from the UCI machine learning repository. The results highlight that the simple input-output layer SNN classifier with a time-varying weight model performs better than the existing algorithms with multiple layers and multiple output neurons. The results of an ablation study on time-varying weight model using JAFFE dataset highlight that using a time-varying weight model instead of a constant weight model significantly improves the performance of the classifier. The performance and the compact architecture of the SEF-M classifier also highlight that the time-varying weight model is computationally powerful than long-term and short-term weight models.   
\section*{Acknowledgement}
This work was supported by the Science and Engineering Research Council of A*STAR (Agency for Science, Technology and Research), Singapore.
\bibliographystyle{ieeetr}
\bibliography{Sefron}

\begin{thebibliography}{10}

\bibitem{Masss_neuron}
W.~Maass, ``{Noisy Spiking Neurons with Temporal Coding have more Computational
  Power than Sigmoidal Neurons},'' {\em Institue of Theoretical Computer
  Science, Technische Universitaet Graz, Austria, Technical Report},
  p.~[Online].Available:http://www.igi.tugraz.at/psfile, 1999.

\bibitem{SpikeProp}
S.~M. Bohte, J.~N. Kok, and H.~{La Poutr{\'{e}}}, ``{Error-backpropagation in
  temporally encoded networks of spiking neurons},'' {\em Neurocomputing},
  vol.~48, pp.~17--37, 2002.

\bibitem{Mutli_SpikeProp}
O.~Booij and H.~Nguyen, ``{A gradient descent rule for spiking neurons emitting
  multiple spikes},'' {\em Information Processing Letters}, vol.~95,
  pp.~552--558, 2005.

\bibitem{STDP_overview}
H.~Markram, W.~Gerstner, and P.~J. Sj{\"{o}}str{\"{o}}m, {\em
  {Spike-Timing-Dependent Plasticity: A Comprehensive Overview}}.
\newblock Frontiers Media SA, 2012.

\bibitem{ReSuMe}
F.~Ponulak and A.~{Kasi{\'{n}}ski }, ``{Supervised Learning in Spiking Neural
  Networks with ReSuMe : Sequence Learning , Classification , and Spike
  Shifting},'' {\em Neural Computation}, vol.~22, no.~2, pp.~467--510, 2010.

\bibitem{SWAT}
J.~J. Wade, L.~J. Mcdaid, J.~A. Santos, and H.~M. Sayers, ``{SWAT : A Spiking
  Neural Network Training Algorithm for Classification Problems},'' {\em IEEE
  transactions on neural networks.}, vol.~21, no.~11, pp.~1817--1830, 2010.

\bibitem{ASA}
X.~Xie, H.~Qu, Z.~Yi, and J.~Kurths, ``{Efficient Training of Supervised
  Spiking Neural Network via Accurate Synaptic-Efficiency Adjustment Method},''
  {\em IEEE transactions on neural networks and learning systems}, vol.~28,
  no.~6, pp.~1411 -- 1424, 2017.

\bibitem{Tempotron}
R.~G{\"{u}}tig and H.~Sompolinsky, ``{The tempotron: a neuron that learns spike
  timing–based decisions},'' {\em Nature Neuroscience}, vol.~9, no.~3,
  pp.~420--428, 2006.

\bibitem{SPAN}
A.~Mohemmed, S.~Schliebs, S.~Matsuda, and N.~Kasabov, ``{Span: Spike Pattern
  Association Neuron for Learning Spatio-Temporal Spike Patterns},'' {\em
  International Journal of Neural Systems}, vol.~22, no.~04, p.~1250012 (17
  pages), 2012.

\bibitem{SRESN}
S.~Dora, K.~Subramanian, S.~Suresh, and N.~Sundararajan, ``{Development of a
  Self-Regulating Evolving Spiking Neural Network for classification
  problem},'' {\em Neurocomputing}, vol.~171, pp.~1216--1229, 2016.

\bibitem{SpikeTemp}
J.~Wang, A.~Belatreche, L.~P. Maguire, and T.~M. Mcginnity, ``{SpikeTemp : An
  Enhanced Rank-Order-Based Learning Approach for Spiking Neural Networks With
  Adaptive Structure},'' {\em IEEE transactions on neural networks and learning
  systems}, vol.~28, no.~1, pp.~30--43, 2017.

\bibitem{Chronotron}
R.~V. Florian, ``{The Chronotron : A Neuron That Learns to Fire Temporally
  Precise Spike Patterns},'' {\em PLoS ONE}, vol.~7, no.~8, p.~e40233, 2012.

\bibitem{OMLA}
S.~Dora, S.~Suresh, and N.~Sundararajan, ``Online meta-neuron based learning
  algorithm for a spiking neural classifier,'' {\em Information Sciences},
  vol.~414, pp.~19 -- 32, 2017.

\bibitem{OSNN}
J.~Wang, A.~Belatreche, L.~Maguire, and T.~M. Mcginnity, ``{An online
  supervised learning method for spiking neural networks with adaptive
  structure},'' {\em Neurocomputing}, vol.~144, pp.~526--536, 2014.

\bibitem{MuSpiNN}
S.~Ghosh-Dastidar and H.~Adeli, ``{A new supervised learning algorithm for
  multiple spiking neural networks with application in epilepsy and seizure
  detection},'' {\em Neural Networks}, vol.~22, no.~10, pp.~1419--1431, 2009.

\bibitem{1996LiawSynapseModel}
J.~S. Liaw and T.~W. Berger, ``{Dynamic synapses: A new concept of neural
  representation and computation},'' {\em Hippocampus}, vol.~6, no.~1996,
  pp.~591--600, 1996.

\bibitem{1996MarkramdepressingSynapse}
M.~V. Tsodyks and H.~Markram, ``{Plasticity of neocortical synapses enables
  transitions between rates and temporal coding},'' {\em Proceedings of ICANN},
  pp.~445--450, 1996.

\bibitem{1997MarkramdepressingSynapse}
M.~Tsodyks and H.~Markram, ``{The neural code between neocortical pyramidal
  neurons depends on neurotransmitter release probability},'' {\em PNAS},
  vol.~94, no.~2, pp.~719--723, 1997.

\bibitem{1998MarkramSynapseModel}
M.~Tsodyks, K.~Pawelzik, and H.~Markram, ``{Neural networks with dynamic
  synapses.},'' {\em Neural computation}, vol.~10, no.~4, pp.~821--835, 1998.

\bibitem{STP_Model2}
L.~F. Abbott, J.~A. Varela, K.~Sen, and S.~B. Nelson, ``{Synaptic Depression
  and Cortical Gain Control},'' {\em Science}, vol.~275, no.~5297,
  pp.~220--224, 1997.

\bibitem{STP_model1}
J.~S. Dittman, A.~C. Kreitzer, and W.~G. Regehr, ``{Interplay between
  facilitation, depression, and residual calcium at three presynaptic
  terminals.},'' {\em The Journal of neuroscience : the official journal of the
  Society for Neuroscience}, vol.~20, no.~4, pp.~1374--1385, 2000.

\bibitem{1999MaassSynapseModel}
W.~Maass and A.~M. Zador, ``{Dynamic stochastic synapses as computational
  units.},'' {\em Neural computation}, vol.~11, no.~4, pp.~903--917, 1999.

\bibitem{2003LiawModelImprovement}
A.~A. Dibazar, H.~H. Namarvar, and T.~W. Berger, ``{A New Approach for Isolated
  word recognition using dynamic synapse neural networks},'' {\em Proceedings
  of the International Joint Conference on Neural Networks}, vol.~4,
  pp.~3146--3150, 2003.

\bibitem{1998LiawModelImprovement}
J.~S. Liaw and T.~W. Berger, ``{Robust speech recognition with dynamic
  synapses},'' {\em 1998 IEEE International Joint Conference on Neural Networks
  Proceedings. IEEE World Congress on Computational Intelligence (Cat.
  No.98CH36227)}, vol.~3, pp.~2175--2179, 1998.

\bibitem{2001LiawmodelImprovement}
H.~H. Namarvar, J.~s. Liaw, and T.~W. Berger, ``{A New Dynamic Synapse Neural
  Network for Speech Recognition},'' {\em Neural Networks, 2001. Proceedings.
  IJCNN '01.}, vol.~4, pp.~2985--2990, 2001.

\bibitem{SEFRON}
A.~Jeyasothy, S.~Sundaram, and N.~Sundararajan, ``Sefron: A new spiking neuron
  model with time-varying synaptic efficacy function for pattern
  classification,'' {\em IEEE Transactions on Neural Networks and Learning
  Systems}, pp.~1--10, 2018.

\bibitem{TMM-SNN}
S.~Dora, S.~Sundaram, and N.~Sundararajan, ``An interclass margin maximization
  learning algorithm for evolving spiking neural network,'' {\em IEEE
  Transactions on Cybernetics}, pp.~1--11, 2018.

\bibitem{UCI}
D.~Dheeru and E.~Karra~Taniskidou, ``{UCI} machine learning repository,'' 2017.

\bibitem{Jaffe}
M.~Lyons, S.~Akamatsu, M.~Kamachi, and J.~Gyoba, ``Coding facial expressions
  with gabor wavelets,'' in {\em Proceedings Third IEEE International
  Conference on Automatic Face and Gesture Recognition}, pp.~200--205, April
  1998.

\end{thebibliography}

\end{document}